\documentclass{isprs} % isprs class modified 23-04-2019 (Dennis Wittich)
\usepackage{subfigure}
\usepackage{setspace}
\usepackage{geometry} % added 27-02-2014 Markus Englich
\usepackage{epstopdf}
\usepackage[labelsep=period]{caption}  % added 14-04-2016 Markus Englich - Recommendation by Sebastian Brocks
\usepackage[british]{babel} 
\usepackage[hang]{footmisc}
\usepackage{amsmath}
\usepackage{breqn}
\usepackage{autobreak}

\begin{document}

\title{Tree-GPT: Modular Large Language Model Expert System for Forest Remote Sensing Image Understanding and Interactive Analysis}
\date{}

% KAO: Remove extra spacing
\author{
 S. Q. Du\textsuperscript{a,b,c}, S.J. Tang\textsuperscript{a,b,c} \thanks{Corresponding author}, W. X. Wang \textsuperscript{a,b,c}, X. M. Li\textsuperscript{a,b,c}, R. Z. Guo\textsuperscript{a,b,c}}

% KAO: Remove extra newline
\address{
	\textsuperscript{a }Research Institute for Smart Cities, School of Architecture and Urban Planning, Shenzhen University, Shenzhen, P.R. China\\
 \textsuperscript{b }{State Key Laboratory of Subtropical Building and Urban Science, Shenzhen, P.R. China}\\
	 \textsuperscript{c }Guangdong–Hong Kong-Macau Joint Laboratory for Smart Cities, P.R. China \\
}

% If the corresponding author is NOT the final author, always add a % space before the subsequent comma, i.e.
% first author name\textsuperscript{a,}\thanks{Corresponding author} , % second author name \textsuperscript{b}, etc.
% thanks to Niclas Borlin 05-05-2016

\commission{3, }{1} %This field is optional. If filled, XX and YY should be replaced by adequate numbers. See https://www2.isprs.org/commissions/
\workinggroup{XX/YY} %This field is optional.
\icwg{}   %This field is optional.

% KAO: Use times symbol
\abstract{

This paper introduces a novel framework, Tree-GPT, which incorporates Large Language Models (LLMs) into the forestry remote sensing data workflow, thereby enhancing the efficiency of data analysis. Currently, LLMs are unable to extract or comprehend information from images and may generate inaccurate text due to a lack of domain knowledge, limiting their use in forestry data analysis. To address this issue, we propose a modular LLM expert system, Tree-GPT, that integrates image understanding modules, domain knowledge bases, and toolchains. This empowers LLMs with the ability to comprehend images, acquire accurate knowledge, generate code, and perform data analysis in a local environment. Specifically, the image understanding module extracts structured information from forest remote sensing images by utilizing automatic or interactive generation of prompts to guide the Segment Anything Model (SAM) in generating and selecting optimal tree segmentation results. The system then calculates tree structural parameters based on these results and stores them in a database. Upon receiving a specific natural language instruction, the LLM generates code based on a thought chain to accomplish the analysis task. The code is then executed by an LLM agent in a local environment and . For ecological parameter calculations, the system retrieves the corresponding knowledge from the knowledge base and inputs it into the LLM to guide the generation of accurate code. We tested this system on several tasks, including Search, Visualization, and Machine Learning Analysis. The prototype system performed well, demonstrating the potential for dynamic usage of LLMs in forestry research and environmental sciences.
}


\maketitle

\section{Introduction}

Analysis of forest remote sensing data is essential for various applications in forestry and environmental sciences\cite{turner2006economic,suratno2009tree,indirabai2019estimation,sun2019characterizing}. The analysis of forest remote sensing data is often subject to temporal constraints, necessitating the exploration of more efficient method. Recent advancements in artificial intelligence have given rise to Large Language Models (LLMs), which are machine learning algorithms capable of comprehending human natural language instruction and generating coherent text \cite{zhao2023survey}. By utilizing natural language as an interface, researchers can employ LLMs to generate data analysis code in a matter of minutes, a task that previously required several tens of minutes to complete. Consequently, the integration of LLMs into the analysis of forest remote sensing data is being considered as a means of enhancing efficiency in a similar manner.

%--------------------------------------------------------------------------------

LMMs like GPT series have left a profound impression due to their powerful intelligence \cite{ouyang2022training}. Naturally, we are curious about whether we can employ similar models to perform intelligent analysis of forest remote sensing data. However, the limited ability of LLMs to comprehend images and generate accurate domain-specific knowledge hinders their application in forest remote sensing data analysis. Vision serves as the primary source of information for human cognition. However, the majority of language models struggle to comprehend visual information, limiting their applicability. In recent years, researchers have investigated methods to enable LLMs to process multimodal information. Among these studies, GPT-4 has emerged as the leading model, exhibiting remarkable performance in understanding and accurately describing natural images \cite{openai2023gpt4}. Additionally, open-source models such as MiniGPT4 have proposed their own solutions \cite{zhu2023minigpt}. However, these approaches exhibit suboptimal performance on remote sensing imagery due to the fact that the majority of these models are trained on natural images. To address this limitation, researchers have proposed multimodal remote sensing datasets and trained models that capable of understanding remote sensing images and performing natural language-based question answering \cite{lobry2020rsvqa,chappuis2022prompt}. These models can understand natural language and provide answers regarding the presence of specific concepts in the images. However, due to their complexity, these model are still not capable of precise quantitative analysis or executing data analysis tasks through code generation. Furthermore, most language models exhibit limited performance when generating text in specific professional domains due to a lack of fine-tuning for those domains. A phenomenon known as “hallucination” is also commonly observed in language models, where they generate information that does not exist in reality, such as providing incorrect interpretations of natural laws \cite{ji2023survey}. Consequently, the lack of remote sensing image understanding ability and accurate professional knowledge are significant issues that restrict the application of general language models in specialized domains.

%————————————————————————————————————————————————————————————————————————————————
To address the first issue and enable LLMs to understand forest remote sensing images, we are considering the integration of a separate image understanding module. This module is built to convert the implicit information contained in the image into explicit, structured information that LLMs can understand. For forest remote sensing data analysis, it typically involves two primary tasks: the individual tree crown (ITC) segmentation  and calculation of tree structural parameters. Among them, the ITC segmentation remains a challenging problem. Current research has explored various methods based on point clouds or images for ITC segmentation. Among them, image-based segmentation methods tend to employ supervised learning algorithms like Mask R-CNN for accurate instance segmentation\cite{beloiu2023individual}. These methods require a large number of high-precision training samples. To reduce the cost of our approach, we are considering whether unsupervised or interactive methods can be used for ITC segmentation to generate structured information that LLMs can comprehend.

Recently, image representation learning \cite{vaswani2017attention,dosovitskiy2020image} and self-supervised learning \cite{he2022masked} have advanced significantly, along with the availability of large-scale data and computational resources \cite{kirillov2023segment}. These developments have enabled the learning of universal features that can be transferred across different domains from massive image collections. In this context, a Large Vision Model named Segment Anything Model (SAM) was proposed \cite{kirillov2023segment}, which can perform zero-shot image segmentation guided by prompts. However, SAM still faces two challenges in individual tree segmentation task: 1) the segmentation quality of SAM is highly sensitive to the input prompt, and the segmentation granularity is not controllable without prior knowledge. 2) SAM generates many redundant masks in its segmentation output, which hinders the identification of individual trees. This paper aims to overcome these challenges and leverage the potential of SAM for single-tree segmentation tasks by proposing a framework that does not require retraining and can be easily adapted to new scenarios.

%————————————————————————————————————————————————————————————————————————————————

To address the second issue of empowering LLMs with domain-specific knowledge and task execution capabilities, we are considering building a specialized knowledge base and an execution agent for LLMs. The specialized knowledge base is a widely used solution to tackle the issue of hallucination in LLMs. Meanwhile, the execution agent acts as a framework that connects LLMs with the local execution environment. LLMs can utilize the execution agent to execute generated code in the local development environment, perform data manipulation, and generate visualization or data analysis results. Currently, there are existing frameworks for knowledge base construction and task execution for LLMs. However, designing the optimal framework is still an open question. For the specific task of forest remote sensing data analysis, it is worth discussing how to adapt the existing frameworks, such as Prompt Engineering, to suit the characteristics of this particular task.

%--------------------------------------------------------------------------------

Following above ideas, we propose a novel framework, Tree-GPT. The purpose of Tree-GPT is to build an expert system capable of understanding forest remote sensing images, possessing domain-specific knowledge in forest ecology, and generating code based on prompts to automatically execute data analysis tasks. The core of this system consists of an LLM as the reasoning engine, an image understanding module to convert image information into text, a Domain Knowledge Base to store professional knowledge for LLM retrieval, and an LLM execution agent for code execution. More specifically:
\begin{itemize}
\item \textbf{LLM:} The LLM and its accompanying backend execution framework are pivotal components of Tree-GPT. In terms of model selection, the study utilizes the cloud-deployed OpenAI GPT-4 model, known for its high performance. However, in theory, an LLM deployed locally would also be suitable.

\item \textbf{Image Understanding:} As the second key module of Tree-GPT, the image understanding module encompasses two functions: tree segmentation and calculation of tree structural parameters. To enable low-cost algorithm transfer, the study employs the Structure-Aware Masking (SAM) as the core and designs a new tree segmentation method to address the challenges associated with SAM. This method starts by employing a pre-trained tree detection model to locate and outline trees (or uses an interactive approach) to generate an initial prompt. Subsequently, the study inputs the initial prompt and a grid prompt into SAM, resulting in redundant tree segmentation outcomes. To refine the segmentation results, a bipartite graph matching model is utilized, leveraging the initial prompt as a guide. After obtaining the segmentation results, additional remote sensing data is incorporated to compute tree structural parameters, which are stored in a relational database for subsequent data analysis access.

\item \textbf{Domain Knowledge Base:} As the third key module of Tree-GPT, the Domain Knowledge Base is a vector database that stores forest ecology domain-specific knowledge in embedded text form. This knowledge base allows for retrieval of crucial information through keyword searches and utilizes it as prompts to guide the LLM in generating accurate professional knowledge outputs.

\item \textbf{LLM Execution Agent:} Lastly, as the execution module of Tree-GPT, the LLM execution agent serves as a backend framework that connects the LLM's code outputs with the local runtime environment. The study has tailored the Prompt Engineering framework to address the specific tasks of forest remote sensing data analysis, including information retrieval, result visualization, and data analysis. Task-oriented prompt design ensures the generation of precise code, facilitating efficient and effective data analysis.
\end{itemize}

\begin{figure*}[htbp]
\centering
\includegraphics{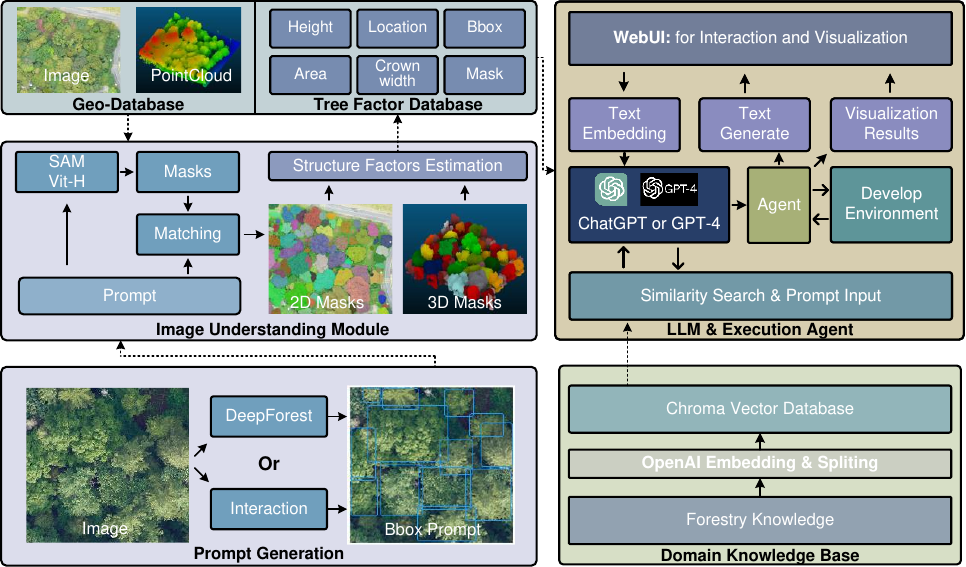}
\caption{\label{fig:Tree-GPT}Overall Framework of the Tree-GPT. }
\end{figure*}

In conclusion, Tree-GPT offers a user-friendly approach that enables real-time data updates and analysis. This reduces the time required for the processing workflow and allows researchers to allocate more resources towards data interpretation and the understanding of ecological patterns. With its understanding and generation capabilities provided by Large Language Models, Tree-GPT offers a pertinent contribution to the field of tree ecological parameter extraction.

\section{Related Works}
\subsection{Individual Tree Crown (ITC) Segmentation}
Efficient and automated ITC segmentation methods are crucial for calculating tree factors \cite{yang2019influence}. Existing methods for ITC segmentation can be classified based on the type of data used, including point cloud based methods and image based method. Current UAV point cloud based methods often utilize unsupervised clustering techniques. Since the highest point of a tree (which can be regarded as the tree’s center) is easily identifiable in the point cloud, these methods typically determine the tree center by leveraging the elevation differences. They then perform point cloud clustering using the center as a seed point to segment the tree contour \cite{strimbu2015graph}. However, accurately describing tree boundaries becomes challenging due to the lack of texture information in point clouds. As a result, most methods rely on strong prior assumptions such as the conical shape assumption to differentiate between points at the boundaries of trees \cite{strimbu2015graph,yang2020individual,qian2023tree}. These methods generally achieve acceptable results in most scenarios. However, in subtropical or tropical regions, trees tend to have broader and more expansive shapes with large crown areas \cite{10.1371/journal.pone.0036131}. As a result, many trees’ growth patterns deviate from the aforementioned prior assumptions, which limits the generalization capability of these methods when applied to new scenarios, necessitating repeated parameter tuning. 

Based on image data, tree segmentation methods primarily utilize the color and texture differences of different trees in RGB images or the tree height differences in the canopy height model (CHM) to segment individual trees. Some studies have employed traditional morphological image segmentation methods, such as watershed algorithms and their variants, to achieve tree segmentation in RGB or CHM images \cite{tochon2015use,wagner2018individual,yang2019influence}. However, the limited utilization of morphological image features hinders the accuracy of these segmentation methods. In recent years, research has focused on deep learning-based single-tree segmentation methods due to the advancement of deep learning techniques. Compared to traditional image segmentation methods, deep learning networks possess stronger feature extraction capabilities and can learn features with certain generalization properties, thus enhancing the generalization of the methods \cite{lecun2015deep}. Currently, most deep learning-based single-tree segmentation methods are based on the Mask-RCNN network \cite{he2017mask,beloiu2023individual}. Some methods have improved edge extraction by enhancing the loss function \cite{zhang2022multi}, while others have considered incorporating multi-modal feature extraction techniques within the Mask-RCNN framework. They simultaneously use texture, color, and elevation information from both RGB and CHM images to improve tree segmentation accuracy \cite{li2022ace}. Although deep learning-based instance segmentation methods for single-tree segmentation show significant improvements over traditional image segmentation methods in terms of robustness and generalization, most supervised deep learning methods suffer from a fundamental limitation: their performance is heavily constrained by the training data. Since deep learning is essentially about finding a set of factors that best fit the distribution of the training samples, the performance of the network is determined by the extent to which the probability distribution of the training samples aligns with the real-world distribution \cite{goodfellow2016deep}. For single-tree segmentation tasks, high-precision open-source tree contour datasets are severely lacking. Moreover, distinguishing and annotating tree contours is extremely costly compared to natural images, making it difficult to rapidly annotate a large amount of data to train a tree segmentation network with strong generalization and zero-shot transfer capability.

\section{Method}\label{sec:MAIN BODY OF TEXT}
Tree-GPT is a specialized system designed to process remote sensing data from forests. It comprises a GPT-4-based LLM, an image understanding module, a Domain Knowledge Base, and an LLM execution agent. The Tree-GPT workflow operates as follows: given a natural language instruction $L$ and a remote sensing image $I$ of the study area, the image understanding module processes $I$ and collaborates with other remote sensing data from the study area to generate tree structure parameters that can be used as inputs for the LLM and for data analysis. These parameters are stored in a relational database $R$. Simultaneously, the professional domain knowledge base module assesses $L$ to determine if it is a knowledge-based question and whether to invoke the professional knowledge base module to answer it. If $L$ is not a knowledge-based question, it proceeds to the LLM execution agent module.

In the LLM execution module, if $L$ does not involve code execution tasks, the output result of the LLM is directly returned in the FastChat frontend. If the result contains code execution tasks, the corresponding code is generated based on prompt templates and executed using the LangChain agent module in a local runtime environment. During execution, data visualization or analysis results are generated by calling image $I$ and database $R$.

\begin{figure*}[h]
\centering
\includegraphics[width=1\textwidth]{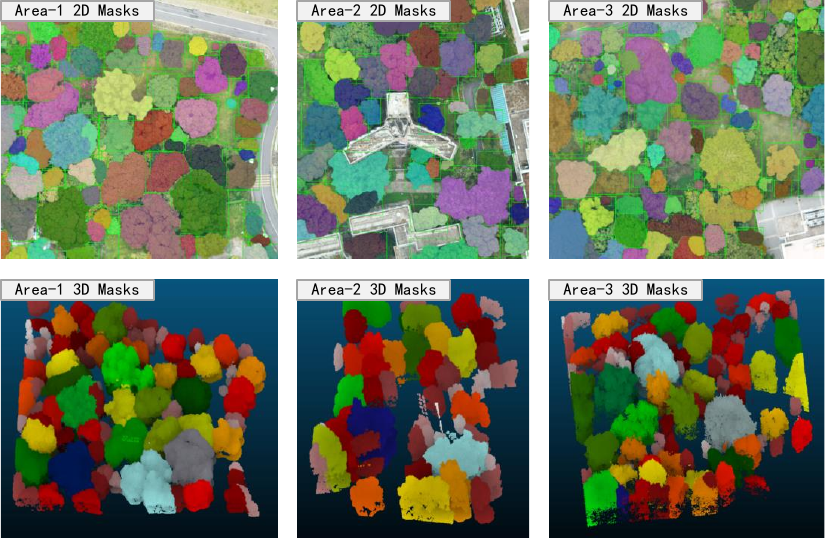}
\caption{\label{fig:Mask}Visualization of individual tree crown(ITC) segmentation results. The first line shows tree crown segmentation results (segmentation in 2D). The second line shows individual tree point cloud segmentation results (3D segmentation by projecting 2D results to point cloud).}
\end{figure*}

\subsection{Image Understanding Module}
\textbf{Mask Generation:}The Tree-GPT model utilizes a tree target detection framework and a bipartite graph matching method to control the segmentation granularity of SAM. This generates tree segmentation results that correspond one-to-one with the tree position. Specifically, the single-tree segmentation module uses points and target frames as SAM’s Prompt at the same time to guide the generation of tree Masks. Among them, the target detection module can use automatic or manual methods to generate tree target detection frames, which are used to guide SAM for tree segmentation and assign semantic information to the segmentation results. The automatic target frame generation method uses RetinaNet \cite{lin2017focal} as the target detection framework with ResNet50 \cite{he2016deep} as the backbone network and DeepForest \cite{weinstein2019individual} pretrained weights. Manual annotation of tree target frames can be used as a supplement to automatic tree target detection to improve the accuracy of tree detection in an interactive manner. In addition to the target frame, we use a 48 $\times$ 48 grid sampling point as the basic Prompt for SAM to deal with situations where BBox Prompt guidance fails. This generates enough Masks for subsequent matching with the target frame and obtains optimal segmentation results of the tree. After obtaining the Prompt, we input Point and BBox Prompt into SAM respectively and merge the two sets of tree segmentation results to obtain redundant tree segmentation results.

\textbf{Best Mask Matching:} After obtaining the redundant tree segmentation results, we consider how to select the optimal tree segmentation results that correspond one-to-one with the real trees. Since SAM’s segmentation results do not contain semantic information, we consider using the automatically or manually obtained tree target detection results as the real position of the tree. We match the redundant Mask with the target detection results to obtain the tree segmentation results corresponding to the tree position. Specifically, we model this matching problem as a bipartite graph matching problem in mathematics. Bipartite graph matching, also known as bipartite matching, is a model in graph theory where given two disjoint sets of vertices A and B, and a set of edges $E$ $\subseteq$ $A$ $\times$ $B$, a matching $M$ is a subset of $E$ such that no two edges share an endpoint. A maximum matching is a matching $M$ of maximum cardinality (maximum number of edges). This problem can be defined as:

\begin{equation} \text{min:} \sum_{i \in A} \sum_{j \in B} c_{ij}x_{ij} \end{equation}
\begin{equation}
    \begin{aligned}
        \text{s.t. } & \sum_{i \in A} x_{ij} = 1, \text{ for all } j \in B \\
        & \sum_{j \in B} x_{ij} = 1, \text{ for all } i \in A \\
        & x_{ij} \in \{0, 1\}, \text{ for all } (i, j) \in E
    \end{aligned}
\end{equation}

Where the matchings are represented by variables $x_{ij}$ for each edge $E(i, j)$. $c_{ij}$ is the cost of matching vertex i to vertex j in the bipartite graph. The constraints ensure that each vertex is matched exactly once.

In short, the bipartite graph matching problem is used to solve a set of elements in set $B$ that correspond one-to-one with elements in set $A$. This correspondence is called maximum matching. When solving bipartite graph matching, it is necessary to define the cost of adding each edge, that is, the Cost matrix. In this article, we aim to maximize the overlap between the segmentation result and our detected target frame. We can use GIoU \cite{rezatofighi2019generalized} to characterize this degree of overlap. The GIoU is defined as IoU minus the area of the smallest enclosing box that covers both bounding boxes P and G, excluding the union area of P and G and can be expressed as follows:
\begin{equation}
IoU = \frac{{\text{{Area of Intersection}}}}{{\text{{Area of Union}}}}
\end{equation}

\begin{equation}
GIoU = IoU - \frac{A_c - A_u}{A_c}
\end{equation}

After obtaining the Cost matrix, we use the classic Hungarian algorithm to solve bipartite graph matching and obtain the optimal segmentation result with the highest overlap with tree position.

\textbf{Tree Structure Factor Estimation:} After obtaining the tree segmentation results, we use the 2D segmentation results to assign labels to the point cloud within the corresponding range. This allows us to obtain the 3D single-tree segmentation results. After obtaining the 3D segmentation, we calculate the tree height, crown width, support height and crown area. We store them in the tree database together with the 2D segmentation results, external target frame and tree position information. Among them, the tree position is stored in the database in pixel coordinates. The 2D tree contour is stored in the database in MS COCO’s \cite{lin2014microsoft} compressed storage format.

\subsection{Domain Knowledge Base}
The Domain Knowledge Base is a vector database that stores embedded text of ecological knowledge in the field of forestry. The module first uses the OpenAIEmbedding API to embed the text of ecological knowledge in the forestry field and convert it into numerical vectors. The text is then segmented using the LangChain framework's text segmentation tool with a block size of 4,000 tokens, and the resulting knowledge vectors are split into several blocks and stored in the Chroma vector database. When Tree-GPT receives a natural language instruction, it is first embedded using the OpenAIEmbedding API and converted into a numerical vector. The module then uses the Facebook AI Similarity Search (FAISS) algorithm to retrieve relevant information from the Chroma database, obtaining retrieval results denoted as SS. These results are ranked, and text blocks with a similarity greater than 0.6 are selected. These selected text blocks are then used as new inputs to the LLM using the prompt template 'Given {context}, could you please explain the meaning of {query}?' to obtain the final inference result. This allows the model to accurately answer knowledge-based questions in the professional domain.

\subsection{LLM Execution Agent}
The LLM Execution Agent is a backend framework in TreeGPT that bridges the LLM's generated code structure with the local runtime environment. Built on the LangChain toolchain, the LLM Execution Agent executes instructions in two steps: Task Planning and Code Execution. In Task Planning, the LLM decomposes natural language instructions into subtasks using prompt templates and generates corresponding content for each subtask. By breaking down tasks, the LLM is more likely to generate accurate code, as the subtasks are simpler. TreeGPT employs task planning templates, as shown in the Table  \ref{Tab:Prompt template}.
\begin{table}[h]
\begin{tabular}{|l|c|c|}
\hline
\multicolumn{1}{|c|}{\textbf{\begin{tabular}[c]{@{}c@{}}Prompt \\ Templete\end{tabular}}} & \textbf{Input} & \textbf{Output} \\ \hline
1. Thought & \begin{tabular}[c]{@{}c@{}}Natural Language \\ Instruction\end{tabular} & \begin{tabular}[c]{@{}c@{}}Task \\ Decomposition\end{tabular} \\ \hline
2. Action & \begin{tabular}[c]{@{}c@{}}Task \\ Decomposition\end{tabular} & \begin{tabular}[c]{@{}c@{}}Type of \\ Sub-Tasks\end{tabular} \\ \hline
3. Action Input & Sub-Tasks & Code \\ \hline
4. Observation & Code & \begin{tabular}[c]{@{}c@{}}Running \\ Results\end{tabular} \\ \hline
5. Thought & \begin{tabular}[c]{@{}c@{}}Running Results + \\ Instruction\end{tabular} & Output Text \\ \hline
6. Final Result & - & \begin{tabular}[c]{@{}c@{}}Output Text+\\ Running Results \\ (Visualization, etc.)\end{tabular} \\ \hline
\end{tabular}
\caption{\label{Tab:Prompt template}Process of LLM Execution Agent.}
\end{table}

\section{Experiment Results}\label{sec:MAIN BODY OF TEXT}

\begin{figure*}[ht]
\centering
\includegraphics[width=1\textwidth]{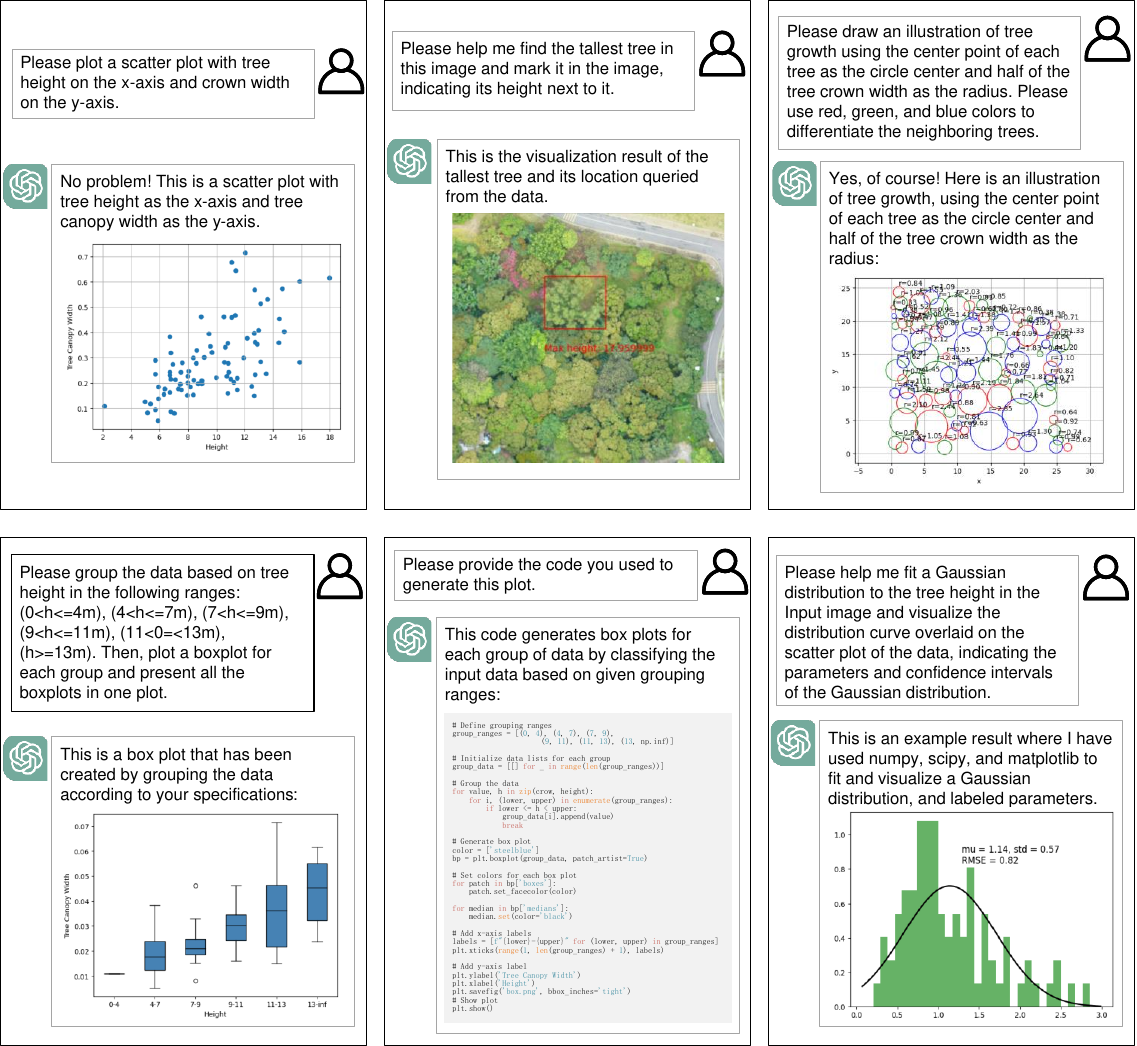}
\caption{\label{fig:ChatResult}The results of Tree-GPT’s operation.}
\end{figure*}

We use data collected from Shenzhen, Guangdong Province, China to verify the effectiveness of our method. The data was collected in 2018, with an RGB orthoimage spatial resolution of 0.025m and a Lidar resolution of 100 points per $m^{2}$. In the experiment, we cropped three 3000$\times$3000 pixel areas to verify the effectiveness of the method. In terms of language model selection, we use the ChatGPT API to access the database to maximize the accuracy of code generation.

Our experiments on Tree-GPT revolve around two tasks: tree segmentation and data visualization and analysis based on natural language. The results are presented in Figure \ref{fig:Mask} and \ref{fig:ChatResult}. As shown in Figure \ref{fig:Mask}, in the tree segmentation task, Tree-GPT’s segmentation results are basically consistent with manual annotations. It can even be said that in some cases the manual annotation results may not be as accurate as SAM’s results. In addition, based on natural language-guided tree parameter query, visualization and analysis results are presented in Figures \ref{fig:ChatResult} respectively. The operating results of Tree-GPT are demonstrated in various tasks, including simple visualization (directly generating scatter plots), information retrieval (finding the tallest tree), complex visualization (drawing a tree growth diagram based on the position and crown width of the tree and generating a box plot of trees grouped by height), code generation, and statistical learning-based analysis and testing of tree ecological parameters (estimating the Gaussian distribution parameters of tree height and using RMSE as a confidence measure). As shown in the figure\ref{fig:ChatResult}, during the query, Tree-GPT can give correct query results and visualize them in the figure. In the data visualization task, Tree-GPT can give basically correct results in the first round. However, it may require multiple manual guidance to achieve the best visualization effect. In the data analysis task, Tree-GPT can implement simple machine learning code for various tasks.

\section{Conclusion}\label{sec:MAIN BODY OF TEXT}
In conclusion, Tree-GPT offers a user-friendly approach that enables real-time data updates and analysis, thereby reducing the time required for the processing workflow and allowing researchers to allocate more resources towards data interpretation and the understanding of ecological patterns. With its understanding and generation capabilities provided by Large Language Models, Tree-GPT offers a pertinent contribution to the field of tree ecological parameter extraction.

\section*{ACKNOWLEDGEMENTS}\label{sec:MAIN BODY OF TEXT}
This work was supported in part by the National Key Research and Development Program of China (Project No. 2022 YFB 3903700), a Shenzhen Science and Technology Program (Project No. JCYJ 20210324093012033), the Natural Science Foundation of Guangdong Province (Project No. 2121A1515012574), the National Natural Science Foundation of China (Project Nos. 71901147, 41901329, 41971354, and 41971341,42001331), Shenzhen Key Laboratory of Digital Twin Technologies for Cities (Project No: ZDSYS20210623101800001).

\bibliography{ref}
\end{document}